# Accelerating Neural Transformer via an Average Attention Network

**Biao Zhang**[1,2], **Deyi Xiong**[3] and **Jinsong Su**[1,2*]
Xiamen University, Xiamen, China 361005[1]
Beijing Advanced Innovation Center for Language Resources[2]
Soochow University, Suzhou, China 215006[3]
zb@stu.xmu.edu.cn, dyxiong@suda.edu.cn, jssu@xmu.edu.cn

## Abstract

With parallelizable attention networks, the neural Transformer is very fast to train. However, due to the auto-regressive architecture and self-attention in the decoder, the decoding procedure becomes slow. To alleviate this issue, we propose an average attention network as an alternative to the self-attention network in the decoder of the neural Transformer. The average attention network consists of two layers, with an average layer that models dependencies on previous positions and a gating layer that is stacked over the average layer to enhance the expressiveness of the proposed attention network. We apply this network on the decoder part of the neural Transformer to replace the original target-side self-attention model. With masking tricks and dynamic programming, our model enables the neural Transformer to decode sentences over four times faster than its original version with almost no loss in training time and translation performance. We conduct a series of experiments on WMT17 translation tasks, where on 6 different language pairs, we obtain robust and consistent speed-ups in decoding.[1]

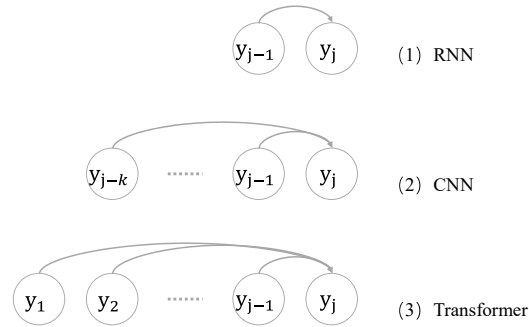

Figure 1: Illustration of the decoding procedure under different neural architectures. We show which previous target words are required to predict the current target word $y_j$ in different NMT architectures. $k$ indicates the filter size of the convolution layer.

## 1 Introduction

The past few years have witnessed the rapid development of neural machine translation (NMT), which translates a source sentence into the target language with an encoder-attention-decoder framework (Sutskever et al., 2014; Bahdanau et al., 2015). Under this framework, various advanced neural architectures have been explored as the backbone network for translation, ranging from recurrent neural networks (RNN) (Sutskever et al., 2014; Luong et al., 2015), convolutional neural networks (CNN) (Gehring et al., 2017a,b) to full attention networks without recurrence and convolution (Vaswani et al., 2017). Particularly, the neural Transformer, relying solely on attention networks, has refreshed state-of-the-art performance on several language pairs (Vaswani et al., 2017).

Most interestingly, the neural Transformer is capable of being fully parallelized at the training phase and modeling intra-/inter-dependencies of source and target sentences within a short path. The parallelization property enables training NMT very quickly, while the dependency modeling property endows the Transformer with strong ability in inducing sentence semantics as well as translation correspondences. However, the decoding of the Transformer cannot enjoy the speed strength of parallelization due to the auto-regressive generation schema in the decoder. And the self-attention

---

*Corresponding author.
[1]Source code is available at https://github.com/bzhangXMU/transformer-aan.

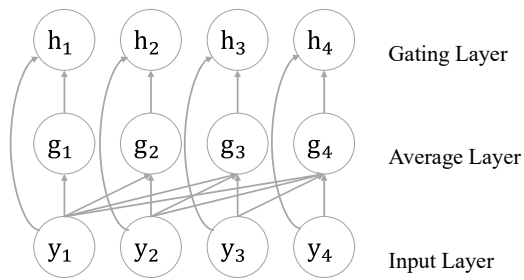

Figure 2: Visualization of the proposed model. For clarity, we show an example with only four words.

network in the decoder even further slows it.

We explain this using Figure 1, where we provide a comparison to RNN- and CNN-based NMT systems. To capture dependencies from previously predicted target words, the self-attention in the neural Transformer requires to calculate adaptive attention weights on all these words (Figure 1 (3)). By contrast, CNN only requires previous $k$ target words (Figure 1 (2)), while RNN merely 1 (Figure 1 (1)). Due to the auto-regressive generation schema, decoding inevitably follows a sequential manner in the Transformer. Therefore the decoding procedure cannot be parallelized. Furthermore, the more target words are generated, the more time the self-attention in the decoder will take to model dependencies. Therefore, preserving the training efficiency of the Transformer on the one hand and accelerating its decoding on the other hand becomes a new and serious challenge.

In this paper, we propose an average attention network (AAN) to handle this challenge. We show the architecture of AAN in Figure 2, which consists of two layers: an *average layer* and *gating layer*. The average layer summarizes history information via a cumulative average operation over previous positions. This is equivalent to a simple attention network where original adaptively computed attention weights are replaced with averaged weights. Upon this layer, we stack a feed forward gating layer to improve the model's expressiveness in describing its inputs.

We use AAN to replace the self-attention part of the neural Transformer's decoder. Considering the characteristic of the cumulative average operation, we develop a masking method to enable parallel computation just like the original self-attention network in the training. In this way, the whole AAN model can be trained totally in parallel so that the training efficiency is ensured. As for the decoding, we can substantially accelerate it by feeding only the previous hidden state to the Transformer decoder just like RNN does. This is achieved with a dynamic programming method.

In spite of its simplicity, our model is capable of modeling complex dependencies. This is because AAN regards each previous word as an equal contributor to current word representation. Therefore, no matter how long the input is, our model can always build up connection signals with previous inputs, which we argue is very crucial for inducing long-range dependencies for machine translation.

We examine our model on WMT17 translation tasks. On 6 different language pairs, our model achieves a speed-up of over 4 times with almost no loss in both translation quality and training speed. In-depth analyses further demonstrate the convergency and advantages of translating long sentences of the proposed AAN.

## 2 Related Work

GRU (Chung et al., 2014) or LSTM (Hochreiter and Schmidhuber, 1997) RNNs are widely used for neural machine translation to deal with long-range dependencies as well as the gradient vanishing issue. A major weakness of RNNs lies at its sequential architecture that completely disables parallel computation. To cope with this problem, Gehring et al. (2017a) propose to use CNN-based encoder as an alternative to RNN, and Gehring et al. (2017b) further develop a completely CNN-based NMT system. However, shallow CNN can only capture local dependencies. Hence, CNN-based NMT normally develops deep archictures to model long-distance dependencies. Different from these studies, Vaswani et al. (2017) propose the Transformer, a neural architecture that abandons recurrence and convolution. It fully relies on attention networks to model translation. The propperties of parallelization and short dependency path significantly improve the training speed as well as model performance for the Transformer. Unfortunately, as we have mentioned in Section 1, it suffers from decoding inefficiency.

The attention mechanism is originally proposed to induce translation-relevant source information for predicting next target word in NMT. It contributes a lot to make NMT outperform SMT. Recently, a variety of efforts are made to further improve its accuracy and capability. Luong et al.

(2015) explore several attention formulations and distinguish local attention from global attention. Zhang et al. (2016) treat RNN as an alternative to the attention to improve model's capability in dealing with long-range dependencies. Yang et al. (2017) introduce a recurrent cycle on the attention layer to enhance the model's memorization of previous translated source words. Zhang et al. (2017a) observe the weak discrimination ability of the attention-generated context vectors and propose a GRU-gated attention network. Kim et al. (2017) further model intrinsic structures inside attention through graphical models. Shen et al. (2017) introduce a direction structure into a self-attention network to integrate both long-range dependencies and temporal order information. Mi et al. (2016) and Liu et al. (2016) employ standard word alignment to supervise the automatically generated attention weights. Our work also focus on the evolution of attention network, but unlike previous work, we seek to simplify the self-attention network so as to accelerate the decoding procedure. The design of our model is partially inspired by the highway network (Srivastava et al., 2015) and the residual network (He et al., 2015).

In the respect of speeding up the decoding of the neural Transformer, Gu et al. (2018) change the auto-regressive architecture to speed up translation by directly generating target words without relying on any previous predictions. However, compared with our work, their model achieves the improvement in decoding speed at the cost of the drop in translation quality. Our model, instead, not only achieves a remarkable gain in terms of decoding speed, but also preserves the translation performance. Developing fast and efficient attention module for the Transformer, to the best of our knowledge, has never been investigated before.

## 3 The Average Attention Network

Given an input layer $\mathbf{y} = \{\mathbf{y}_1, \mathbf{y}_2, \ldots, \mathbf{y}_m\}$, AAN first employs a cumulative-average operation to generate context-sensitive representation for each input embedding as follows (Figure 2 *Average Layer*):

$$\mathbf{g}_j = \text{FFN}\left(\frac{1}{j}\sum_{k=1}^{j}\mathbf{y}_k\right) \quad (1)$$

where FFN $(\cdot)$ denotes the position-wise feed-forward network proposed by Vaswani et al. (2017), and both $\mathbf{y}_k$ and $\mathbf{g}_j$ have a dimensionality of $d$. Intuitively, AAN replaces the original dynamically computed attention weights by the self-attention network in the decoder of the neural Transformer with simple and fixed average weights ($\frac{1}{j}$). In spite of its simplicity, the cumulative-average operation is very crucial for AAN because it builds up dependencies with previous input embeddings so that the generated representations are not independent of each other. Another benefit from the cumulative-average operation is that no matter how long the input is, the connection strength with each previous input embedding is invariant, which ensures the capability of AAN in modeling long-range dependencies.

We treat $\mathbf{g}_j$ as a contextual representation for the $j$-th input, and apply a feed-forward gating layer upon it as well as $\mathbf{y}_j$ to enrich the non-linear expressiveness of AAN:

$$\begin{aligned}\mathbf{i}_j, \mathbf{f}_j &= \sigma\left(\mathbf{W}\left[\mathbf{y}_j; \mathbf{g}_j\right]\right) \\ \tilde{\mathbf{h}}_j &= \mathbf{i}_j \odot \mathbf{y}_j + \mathbf{f}_j \odot \mathbf{g}_j\end{aligned} \quad (2)$$

where $[\cdot; \cdot]$ denotes concatenation operation, and $\odot$ indicates element-wise multiplication. $\mathbf{i}_j$ and $\mathbf{f}_j$ are the *input* and *forget* gate respectively. Via this gating layer, AAN can control how much past information can be preserved from previous context $\mathbf{g}_j$ and how much new information can be captured from current input $\mathbf{y}_j$. This helps our model to detect correlations inside input embeddings.

Following the architecture design in the neural Transformer (Vaswani et al., 2017), we employ a residual connection between the input layer and gating layer, followed by layer normalization to stabilize the scale of both output and gradient:

$$\mathbf{h}_j = \text{LayerNorm}\left(\mathbf{y}_j + \tilde{\mathbf{h}}_j\right) \quad (3)$$

We refer to the whole procedure formulated in Eq. (1∼3) as original AAN $(\cdot)$ in following sections.

### 3.1 Parallelization in Training

A computation bottleneck of the original AAN described above is that the cumulative-average operation in Eq. (1) can only be performed sequentially. That is, this operation can not be parallelized. Fortunately, as the average is not a complex computation, we can use a masking trick to enable full parallelization of this operation.

We show the masking trick in Figure 3, where input embeddings are directly converted into their corresponding cumulative-averaged outputs

| Model | Complexity | Sequential Operations | Maximum Path Length |
|---|---|---|---|
| Self-attention | $\mathcal{O}\left(n^2 \cdot d + n \cdot d^2\right)$ | $\mathcal{O}\left(1\right)$ | $\mathcal{O}\left(1\right)$ |
| Original AAN | $\mathcal{O}\left(n \cdot d^2\right)$ | $\mathcal{O}\left(n\right)$ | $\mathcal{O}\left(1\right)$ |
| Masked AAN | $\mathcal{O}\left(n^2 \cdot d + n \cdot d^2\right)$ | $\mathcal{O}\left(1\right)$ | $\mathcal{O}\left(1\right)$ |

Table 1: Maximum path lengths, model complexity and minimum number of sequential operations for different models. $n$ is the sentence length and $d$ is the representation dimension.

$$\begin{pmatrix} 1 & 0 & 0 & 0 \\ 1/2 & 1/2 & 0 & 0 \\ 1/3 & 1/3 & 1/3 & 0 \\ 1/4 & 1/4 & 1/4 & 1/4 \end{pmatrix} \times \begin{pmatrix} y_1 \\ y_2 \\ y_3 \\ y_4 \end{pmatrix} = \begin{pmatrix} y_1 \\ \frac{y_1 + y_2}{2} \\ \frac{y_1 + y_2 + y_3}{3} \\ \frac{y_1 + y_2 + y_3 + y_4}{4} \end{pmatrix}$$

Mask Matrix

Figure 3: Visualization of parallel implementation for the cumulative-average operation enabled by a mask matrix. $\{\mathbf{y}_1, \mathbf{y}_2, \mathbf{y}_3, \mathbf{y}_4\}$ are the input embeddings.

through a masking matrix. In this way, all the components inside $\text{AAN}\left(\cdot\right)$ can enjoy full parallelization, assuring its computational efficiency. We refer to this AAN as masked AAN.

### 3.2 Model Analysis

In this section, we provide a thorough analysis for AAN in comparison to the original self-attention model used by Vaswani et al. (2017). Unlike our AAN, the self-attention model leverages a scaled dot-product function rather than the average operation to compute attention weights:

$$\mathbf{Q}, \mathbf{K}, \mathbf{V} = f\left(\mathbf{Y}\right)$$
$$\text{Self-Attention}\left(\mathbf{Q}, \mathbf{K}, \mathbf{V}\right) = \text{softmax}\left(\frac{\mathbf{Q}\mathbf{K}^T}{\sqrt{d}}\right)\mathbf{V} \quad (4)$$

where $\mathbf{Y} \in \mathbb{R}^{n \times d}$ is the input matrix, $f\left(\cdot\right)$ is a mapping function and $\mathbf{Q}, \mathbf{K}, \mathbf{V} \in \mathbb{R}^{n \times d}$ are the corresponding queries, keys and values. Following Vaswani et al. (2017), we compare both models in terms of computational complexity, minimum number of sequential operations required and maximum path length that a dependency signal between any two positions has to traverse in the network. Table 1 summarizes the comparison results.

Our AAN has a maximum path length of $\mathcal{O}\left(1\right)$, because it can directly capture dependencies between any two input embeddings. For the original AAN, the nature of its sequential computation enlarges its minimum number sequential operations to $\mathcal{O}\left(n\right)$. However, due to its lack of position-wise masked projection, it only consumes a computational complexity of $\mathcal{O}\left(n \cdot d^2\right)$. By contrast, both self-attention and masked AAN have a computational complexity of $\mathcal{O}\left(n^2 \cdot d + n \cdot d^2\right)$, and require only $\mathcal{O}\left(1\right)$ sequential operation. Theoretically, our masked AAN performs very similarly to the self-attention according to Table 1. We therefore use the masked version of AAN during training throughout all our experiments.

### 3.3 Decoding Acceleration

Differing noticeably from the self-attention in the Transformer, our AAN can be accelerated in the decoding phase via dynamic programming thanks to the simple average calculation. Particularly, we can decompose Eq. (1) into the following two steps:

$$\tilde{\mathbf{g}}_j = \tilde{\mathbf{g}}_{j-1} + \mathbf{y}_j \quad (5)$$
$$\mathbf{g}_j = \text{FFN}\left(\frac{\tilde{\mathbf{g}}_j}{j}\right) \quad (6)$$

where $\tilde{\mathbf{g}}_0 = \mathbf{0}$. In doing so, our model can compute the $j$-th input representation based on only one previous state $\tilde{\mathbf{g}}_{j-1}$, instead of relying on all previous states as the self-attention does. In this way, our model can be substantially accelerated during the decoding phase.

## 4 Neural Transformer with AAN

The neural Transformer models translation through an encoder-decoder framework, with each layer involving an attention network followed by a feed forward network (Vaswani et al., 2017). We apply our masked AAN to replace the self-attention network in its decoder part, and illustrate the overall architecture in Figure 4.

Given a source sentence $\mathbf{x} = \{\mathbf{x}_1, \mathbf{x}_2, \ldots, \mathbf{x}_n\}$, the Transformer leverages its encoder to induce source-side semantics and dependencies so as to enable its decoder to recover the encoded information in a target language. The encoder is composed of a stack of $N = 6$ identical layers, each of which has two sub-layers:

$$\tilde{\mathbf{h}}^l = \text{LayerNorm}\left(\mathbf{h}^{l-1} + \text{MHAtt}\left(\mathbf{h}^{l-1}, \mathbf{h}^{l-1}\right)\right)$$
$$\mathbf{h}^l = \text{LayerNorm}\left(\tilde{\mathbf{h}}^l + \text{FFN}\left(\tilde{\mathbf{h}}^l\right)\right) \quad (7)$$

where the superscript $l$ indicates layer depth, and MHAtt denotes the multi-head attention mechanism proposed by Vaswani et al. (2017).

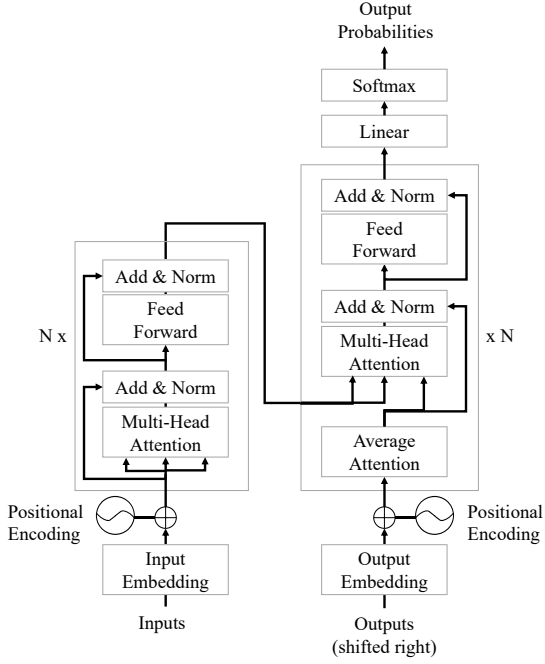

Figure 4: The new Transformer architecture with the proposed average attention network.

Based on the encoded source representation $\mathbf{h}^N$, the Transformer relies on its decoder to generate corresponding target translation $\mathbf{y} = \{\mathbf{y}_1, \mathbf{y}_2, \ldots, \mathbf{y}_m\}$. Similar to the encoder, the decoder also consists of a stack of $N = 6$ identical layers. For each layer in our architecture, the first sub-layer is our proposed average attention network, aiming at capturing target-side dependencies with previous predicted words:

$$\tilde{\mathbf{s}}^l = \text{AAN}\left(\mathbf{s}^{l-1}\right) \quad (8)$$

Carrying these dependencies, the decoder stacks another two sub-layers to seek translation-relevant source semantics for bridging the gap between the source and target language:

$$\mathbf{s}_c^l = \text{LayerNorm}\left(\tilde{\mathbf{s}}^l + \text{MHAtt}\left(\tilde{\mathbf{s}}^l, \mathbf{h}^N\right)\right)$$
$$\mathbf{s}^l = \text{LayerNorm}\left(\mathbf{s}_c^l + \text{FFN}\left(\mathbf{s}_c^l\right)\right) \quad (9)$$

We use subscript $c$ to denote the source-informed target representation. Upon the top layer of this decoder, translation is performed where a linear transformation and softmax activation are applied to compute the probability of the next token based on $\mathbf{s}^N$

To memorize position information, the Transformer augments its input layer $\mathbf{h}^0 = \mathbf{x}, \mathbf{s}^0 = \mathbf{y}$ with frequency-based positional encodings. The whole model is a large, single neural network, and can be trained on a large-scale bilingual corpus with a maximum likelihood objective. We refer readers to (Vaswani et al., 2017) for more details.

## 5 Experiments

### 5.1 WMT14 English-German Translation

We examine various aspects of our AAN on this translation task. The training data consist of 4.5M sentence pairs, involving about 116M English words and 110M German words. We used newstest2013 as the development set for model selection, and newstest2014 as the test set. We evaluated translation quality via case-sensitive BLEU metric (Papineni et al., 2002).

#### 5.1.1 Model Settings

We applied byte pair encoding algorithm (Sennrich et al., 2016) to encode all sentences and limited the vocabulary size to 32K. All out-of-vocabulary words were mapped to an unique token "*unk*". We set the dimensionality $d$ of all input and output layers to 512, and that of inner-FFN layer to 2048. We employed 8 parallel attention heads in both encoder and decoder layers. We batched sentence pairs together so that they were approximately of the same length, and each batch had roughly 25000 source and target tokens. During training, we used label smoothing with value $\epsilon_{ls} = 0.1$, attention dropout and residual dropout with a rate of $p = 0.1$. During decoding, we employed beam search algorithm and set the beam size to 4. Adam optimizer (Kingma and Ba, 2015) with $\beta_1 = 0.9, \beta_2 = 0.98$ and $\epsilon = 10^{-9}$ was used to tune model parameters, and the learning rate was varied under a warm-up strategy with $warmup\_steps = 4000$ (Vaswani et al., 2017).

| Model | BLEU |
|---|---|
| Transformer | 26.37 |
| Our Model | 26.31 |
| Our Model w/o FFN | 26.05 |
| Our Model w/o Gate | 25.91 |

Table 2: Case-sensitive tokenized BLEU score on WMT14 English-German translation. BLEU scores are calculated using *multi-bleu.perl*.

The maximum number of training steps was set to 100K. Weights of target-side embedding and output weight matrix were tied for all models. We implemented our model with masking tricks based on the open-sourced *thumt* (Zhang et al., 2017b)[2], and trained and evaluated all models on a single NVIDIA GeForce GTX 1080 GPU. For evaluation, we averaged last five models saved with an interval of 1500 training steps.

### 5.1.2 Translation Performance

Table 2 reports the translation results. On the same dataset, the Transformer yields a BLEU score of 26.37, while our model achieves 26.31. Both results are almost the same with no significant difference. Clearly, our model is capable of capturing complex translation correspondences so as to generate high-quality translations as effective as the Transformer.

We also show an ablation study in terms of the FFN($\cdot$) network in Eq. (1) and the gating layer in Eq. (2). Table 2 shows that without the FFN network, the performance of our model drops 0.26 BLEU points. This degeneration is enlarged to 0.40 BLEU points when the gating layer is not available. In order to reach comparable performance with the original Transformer, integrating both components is desired.

### 5.1.3 Analysis on Convergency

Different neural architectures might require different number of training steps to converge. In this section, we testify whether our AAN would reveal different characteristics with respect to convergency. We show the loss curve of both the Transformer and our model in Figure 5.

Surprisingly, both model show highly similar tendency, and successfully converge in the end. To train a high-quality translation system, our model consumes almost the same number of training steps as the Transformer. This strongly suggests that replacing the self-attention network with our AAN does not have negative impact on the convergency of the entire model.

[2]https://github.com/thumt/THUMT

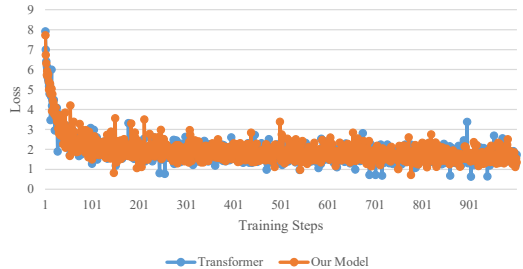

Figure 5: Convergence visualization. The horizontal axis denotes training steps scaled by $10^2$, and the vertical axis indicates training loss. Roughly, our model converges similarly to the Transformer.

|  | Transformer | Our Model | $\triangle_r$ |
|---|---|---|---|
| *Training* | 0.2474 | 0.2464 | 1.00 |
| *Decoding* | | | |
| *beam=4* | 0.1804 | 0.0488 | 3.70 |
| *beam=8* | 0.3576 | 0.0881 | 4.06 |
| *beam=12* | 0.5503 | 0.1291 | 4.26 |
| *beam=16* | 0.7323 | 0.1700 | 4.31 |
| *beam=20* | 0.9172 | 0.2122 | 4.32 |

Table 3: Time required for training and decoding. *Training* denotes the number of global training steps processed per second; *Decoding* indicates the amount of time in seconds required for translating one sentence, which is averaged over the whole newstest2014 dataset. $\triangle_r$ shows the ratio between the Transformer and our model.

### 5.1.4 Analysis on Speed

In Section 3, we demonstrate in theory that our AAN is as efficient as the self-attention during training, but can be substantially accelerated during decoding. In this section, we provide quantitative evidences to examine this point.

We show the training and decoding speed of both the Transformer and our model in Table 3. During training, our model performs approximately 0.2464 training steps per second, while the Transformer processes around 0.2474. This indicates that our model shares similar computational strengths with the Transformer during training, which resonates with the computational analysis in Section 3.

When it comes to decoding procedure, the time of our model required to translate one sentence

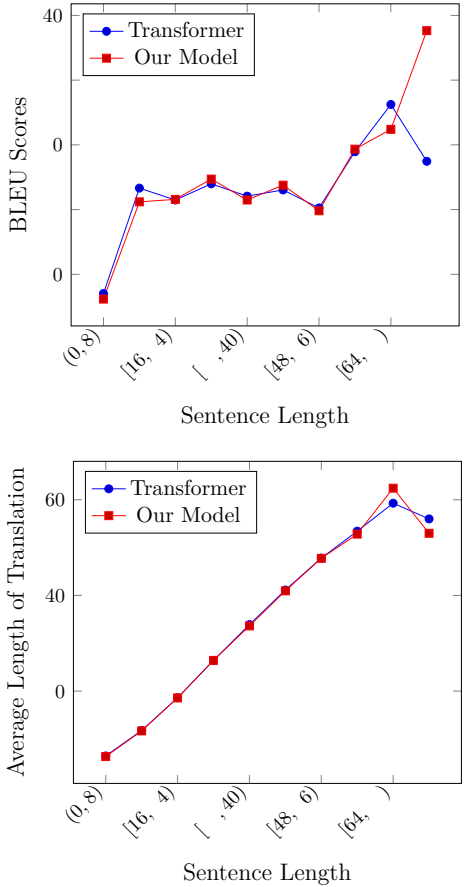

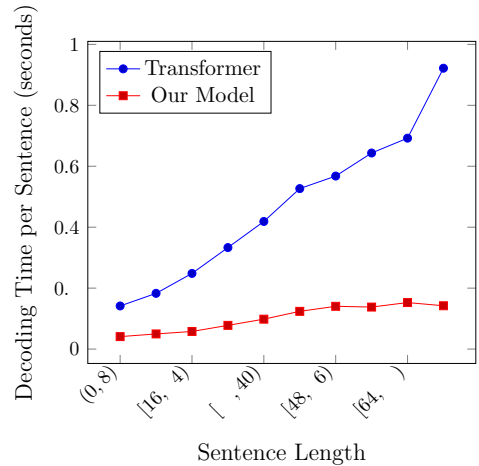

Figure 7: Average time required for translating one source sentence vs. the length of the source sentence. With the increase of sentence length, our model shows more clear and significant advantage over the Transformer in terms of the decoding speed.

Figure 6: Translation statistics on WMT14 English-German test set (newstest14) with respect to the length of source sentences. The top figure shows tokenized BLEU score, and the bottom one shows the average length of translations, both visa-vis sentence length

is only a quarter of that of the Transformer, with beam size ranging from 4 to 20. Another noticeable feature is that as the beam size increases, the ratio of required decoding time between the Transformer and our model is consistently enlarged. This demonstrates empirically that our model, enhanced with the dynamic decoding acceleration algorithm (Section 3.3), can significantly improve the decoding speed of the Transformer.

### 5.1.5 Effects on Sentence Length

A serious common challenge for NMT is to translate long source sentences as handling long-distance dependencies and under-translation issues becomes more difficult for longer sentences. Our proposed AAN uses simple cumulative-average operations to deal with long-range dependencies. We want to examine the effectiveness of these operations on long sentence translation. For this, we provide the translation results along sentence length in Figure 6.

We find that both the Transformer and our model generate very similar translations in terms of BLEU score and translation length, and obtain rather promising performance on long source sentences. More specifically, our model yields relatively shorter translation length on the longest source sentences but significantly better translation quality. This suggests that in spite of the simplicity of the cumulative-average operations, our AAN can indeed capture long-range dependences desired for translating long source sentences.

Generally, the decoder takes more time for translating longer sentences. When it comes to the Transformer, this time issue of translating long sentences becomes notably severe as all previous predicted words must be included for estimating both self-attention weights and word prediction. We show the average time required for translating a source sentence with respect to its sentence length in Figure 7. Obviously, the decoding time of the Transformer grows dramatically with the increase of sentence length, while that of our model rises rather slowly. We contribute this great decoding advantage of our model over the Transformer to the average attention architecture which enables our model to perform next-word prediction by calculating information just from the previous hidden state, rather than considering all previous inputs like the self-attention in the Transformer's decoder.

|  | Case-sensitive BLEU | | | | Case-insensitive BLEU | | | |
| --- | --- | --- | --- | --- | --- | --- | --- | --- |
|  | winner | Transformer | Our Model | $\triangle_d$ | winner | Transformer | Our Model | $\triangle_d$ |
| En→De | 28.3 | 27.33 | 27.22 | -0.11 | 28.9 | 27.92 | 27.80 | -0.12 |
| De→En | 35.1 | 32.63 | 32.73 | +0.10 | 36.5 | 34.06 | 34.13 | +0.07 |
| En→Fi | 20.7 | 21.00 | 20.87 | -0.13 | 21.1 | 21.54 | 21.47 | -0.07 |
| Fi→En | 20.5 | 25.19 | 24.78 | -0.41 | 21.4 | 26.22 | 25.74 | -0.48 |
| En→Lv | 21.1 | 16.83 | 16.63 | -0.20 | 21.6 | 17.42 | 17.23 | -0.19 |
| Lv→En | 21.9 | 17.57 | 17.51 | -0.06 | 22.9 | 18.48 | 18.30 | -0.18 |
| En→Ru | 29.8 | 27.82 | 27.73 | -0.09 | 29.8 | 27.83 | 27.74 | -0.09 |
| Ru→En | 34.7 | 31.51 | 31.36 | -0.15 | 35.6 | 32.59 | 32.36 | -0.23 |
| En→Tr | 18.1 | 12.11 | 11.59 | -0.52 | 18.4 | 12.56 | 12.03 | -0.53 |
| Tr→En | 20.1 | 16.19 | 15.84 | -0.35 | 20.9 | 16.93 | 16.57 | -0.36 |
| En→Cs | 23.5 | 21.53 | 21.12 | -0.41 | 24.1 | 22.07 | 21.66 | -0.41 |
| Cs→En | 30.9 | 27.49 | 27.45 | -0.04 | 31.9 | 28.41 | 28.33 | -0.08 |

Table 4: Detokenized BLEU scores for WMT17 translation tasks. Results are reported with *multi-bleu-detok.perl*. "*winner*" denotes the translation results generated by the WMT17 winning systems. $\triangle_d$ indicates the difference between our model and the Transformer.

### 5.2 WMT17 Translation Tasks

We further demonstrate the effectiveness of our model on six WMT17 translation tasks in both directions (12 translation directions in total). These tasks contain the following language pairs:

- **En-De**: The English-German language pair. This training corpus consists of 5.85M sentence pairs, with 141M English words and 135M German words. We used the concatenation of newstest2014, newstest2015 and newstest2016 as the development set, and the newstest2017 as the test set.

- **En-Fi**: The English-Finnish language pair. This training corpus consists of 2.63M sentence pairs, with 63M English words and 45M Finnish words. We used the concatenation of newstest2015, newsdev2015, newstest2016 and newstestB2016 as the development set, and the newstest2017 as the test set.

- **En-Lv**: The English-Latvian language pair. This training corpus consists of 4.46M sentence pairs, with 63M English words and 52M Latvian words. We used the newsdev2017 as the development set, and the newstest2017 as the test set.

- **En-Ru**: The English-Russian language pair. This training corpus consists of 25M sentence pairs, with 601M English words and 567M Russian words. We used the concatenation of newstest2014, newstest2015 and newstest2016 as the development set, and the newstest2017 as the test set.

- **En-Tr**: The English-Turkish language pair. This training corpus consists of 0.21M sentence pairs, with 5.2M English words and 4.6M Turkish words. We used the concatenation of newsdev2016 and newstest2016 as the development set, and the newstest2017 as the test set.

- **En-Cs**: The English-Czech language pair. This training corpus consists of 52M sentence pairs, with 674M English words and 571M Czech words. We used the concatenation of newstest2014, newstest2015 and newstest2016 as the development set, and the newstest2017 as the test set.

Interestingly, these translation tasks involves training corpora with different scales (ranging from 0.21M to 52M sentence pairs). This help us thoroughly examine the ability of our model on different sizes of training data. All these preprocessed datasets are publicly available, and can be downloaded from WMT17 official website.[3]

We used the same modeling settings as in the WMT14 English-German translation task except for the number of training steps for En-Fi and En-Tr, which we set to 60K and 10K respectively. In addition, to compare with official results, we reported both case-sensitive and case-insensitive detokenized BLEU scores.

---
[3] http://data.statmt.org/wmt17/translation-task/preprocessed/

|  | Transformer | Our Model | $\triangle_r$ |
|---|---|---|---|
| En→De | 0.1411 | 0.02871 | 4.91 |
| De→En | 0.1255 | 0.02422 | 5.18 |
| En→Fi | 0.1289 | 0.02423 | 5.32 |
| Fi→En | 0.1285 | 0.02336 | 5.50 |
| En→Lv | 0.1850 | 0.03167 | 5.84 |
| Lv→En | 0.1980 | 0.03123 | 6.34 |
| En→Ru | 0.1821 | 0.03140 | 5.80 |
| Ru→En | 0.1595 | 0.02778 | 5.74 |
| En→Tr | 0.2078 | 0.02968 | 7.00 |
| Tr→En | 0.1886 | 0.03027 | 6.23 |
| En→Cs | 0.1150 | 0.02425 | 4.74 |
| Cs→En | 0.1178 | 0.02659 | 4.43 |

Table 5: Average seconds required for decoding one source sentence on WMT17 translation tasks.

### 5.2.1 Translation Results

Table 4 shows the overall results on 12 translation directions. We also provide the results from WMT17 winning systems[4]. Notice that unlike the Transformer and our model, these winner systems typically use model ensemble, system combination and large-scale monolingual corpus.

Although different languages have different linguistic and syntactic structures, our model consistently yields rather competitive results against the Transformer on all language pairs in both directions. Particularly, on the De→En translation task, our model achieves a slight improvement of 0.10/0.07 case-sensitive/case-insensitive BLEU points over the Transformer. The largest performance gap between our model and the Transformer occurs on the En→Tr translation task, where our model is lower than the Transformer by 0.52/0.53 case-sensitive/case-insensitive BLEU points. We conjecture that this difference may be due to the small training corpus of the En-Tr task. In all, these results suggest that our AAN is able to perform comparably to Transformer on different language pairs with different scales of training data.

We also show the decoding speed of both the Transformer and our model in Table 5. On all languages in both directions, our model yields significant and consistent improvements over the Transformer in terms of decoding speed. Our model decodes more than 4 times faster than the Transformer. Surprisingly, our model just consumes 0.02968 seconds to translate one source sentence on the En→Tr language pair, only a seventh of the decoding time of the Transformer. These results show that the benefit of decoding acceleration from the proposed average attention structure is language-invariant, and can be easily adapted to other translation tasks.

## 6 Conclusion and Future Work

In this paper, we have described the average attention network that considerably alleviates the decoding bottleneck of the neural Transformer. Our model employs a cumulative average operation to capture important contextual clues from previous target words, and a feed forward gating layer to enrich the expressiveness of learned hidden representations. The model is further enhanced with a masking trick and a dynamic programming method to accelerate the Transformer's decoder. Extensive experiments on one WMT14 and six WMT17 language pairs demonstrate that the proposed average attention network is able to speed up the Transformer's decoder by over 4 times.

In the future, we plan to apply our model on other sequence to sequence learning tasks. We will also attempt to improve our model to enhance its modeling ability so as to consistently outperform the original neural Transformer.

## 7 Acknowledgments

The authors were supported by Beijing Advanced Innovation Center for Language Resources, National Natural Science Foundation of China (Nos. 61672440 and 61622209), the Fundamental Research Funds for the Central Universities (Grant No. ZK1024), and Scientific Research Project of National Language Committee of China (Grant No. YB135-49). Biao Zhang greatly acknowledges the support of the Baidu Scholarship. We also thank the reviewers for their insightful comments.

---
[4]http://matrix.statmt.org/matrix